\documentclass{article} 
\usepackage[preprint]{neurips_2023}

\usepackage{amsmath,amsfonts,bm}

\def\eqref#1{equation~\ref{#1}}

\def\1{\bm{1}}

\def\ra{{\textnormal{a}}}

\def\rx{{\textnormal{x}}}

\def\rva{{\mathbf{a}}}

\def\erva{{\textnormal{a}}}

\def\ervx{{\textnormal{x}}}

\def\rmA{{\mathbf{A}}}

\def\vmu{{\bm{\mu}}}
\def\vtheta{{\bm{\theta}}}
\def\va{{\bm{a}}}

\def\ve{{\bm{e}}}

\def\vx{{\bm{x}}}

\def\eva{{a}}

\def\mA{{\bm{A}}}

\def\mH{{\bm{H}}}
\def\mI{{\bm{I}}}
\def\mJ{{\bm{J}}}

\def\mX{{\bm{X}}}

\def\mSigma{{\bm{\Sigma}}}

\DeclareMathAlphabet{\mathsfit}{\encodingdefault}{\sfdefault}{m}{sl}
\SetMathAlphabet{\mathsfit}{bold}{\encodingdefault}{\sfdefault}{bx}{n}
\newcommand{\tens}[1]{\bm{\mathsfit{#1}}}
\def\tA{{\tens{A}}}

\def\tX{{\tens{X}}}

\def\gG{{\mathcal{G}}}

\def\sA{{\mathbb{A}}}
\def\sB{{\mathbb{B}}}

\def\sS{{\mathbb{S}}}

\def\emA{{A}}

\newcommand{\etens}[1]{\mathsfit{#1}}

\def\etA{{\etens{A}}}

\newcommand{\E}{\mathbb{E}}

\newcommand{\R}{\mathbb{R}}

\newcommand{\KL}{D_{\mathrm{KL}}}
\newcommand{\Var}{\mathrm{Var}}

\newcommand{\Cov}{\mathrm{Cov}}

\newcommand{\normltwo}{L^2}
\newcommand{\normlp}{L^p}

\newcommand{\parents}{Pa} 

\usepackage{hyperref}
\usepackage{url}

\usepackage{graphicx} 
\usepackage{amsmath,amsthm,amssymb,amsfonts}
\usepackage[utf8]{inputenc} 
\usepackage{multirow}
\usepackage[textwidth=3.2cm, size=small]{todonotes}
\usepackage[T1]{fontenc}    
\usepackage{hyperref}       
\usepackage{url}            
\usepackage{booktabs}       
\usepackage{nicefrac}       
\usepackage{microtype}      
\usepackage{xcolor}         
\usepackage{bm}
\usepackage{pifont}
\usepackage[normalem]{ulem}
\usepackage{soul}
\usepackage{mathtools}
\usepackage[inline]{enumitem}
\usepackage{tikz}
\usepackage{graphicx}
\usepackage{natbib}
\usepackage{pdflscape}
\usepackage[makeroom]{cancel}
\usepackage{wrapfig}

\newcommand{\ii}{\text{i}}

\usepackage{tikz}
\usetikzlibrary{graphs}

\usepackage{physics} 

\newtheorem{theorem}{Theorem}

\newtheorem{corollary}{Corollary}
\newtheorem{proposition}{Proposition}

\newtheorem{definition}{Definition}
\usepackage{comment}
\usepackage{tikz}
\usepackage{pgfplots}
\usepackage{soul}
\definecolor{col1}{HTML}{FE6100}
\definecolor{col2}{HTML}{DC267F}
\definecolor{col3}{HTML}{785EF0}
\definecolor{col4}{HTML}{648FFF}
\usepackage{xcolor}

\usepackage{clipboard}
\newcommand{\rsqrt}[1]{#1^{-\frac{1}{2}}}
\renewcommand{\sqrt}[1]{#1^{\frac{1}{2}}}

\newcommand{\red}[1]{\textcolor{red}{#1}}

\newcommand{\rete}{DLGNet}
\newcommand{\reteLONG}{Directed Line Graph Network}

\newcommand{\laplaciano}{Directed Line Graph Laplacian}

\usepackage[greek,british]{babel}
\usepackage{teubner}

\usepackage{scalerel,amssymb}

\usepackage{array}
\newcolumntype{H}{>{\setbox0=\hbox\bgroup}c<{\egroup}@{}}
\usepackage{hyperref}
\usepackage{url}

\renewcommand{\red}[1]{\textcolor{black}#1}

\title{\rete{}: Hyperedge Classification through Directed Line Graphs for Chemical Reactions}

\author{%
  Stefano Fiorini\\
  Pattern Analysis \& Computer Vision \\
  Istituto Italiano di Tecnologia \\
  Genova, Italy \\
  \texttt{stefano.fiorini@iit.it}
  \And
  Giulia M. Bovolenta \\
  Atomistic Simulations \\
  Istituto Italiano di Tecnologia\\
  Genova, Italy \\
  \And
  Stefano Coniglio \\
  Department of Economics\\
  University of Bergamo\\
  Bergamo, Italy \\
  \And
  Michele Ciavotta \\
  University of Milano-Bicocca \\
  Department of Informatics, \\
  Systems and Communication\\
  Milano, Italy \\
  \And
  Pietro Morerio\\
  Pattern Analysis \& Computer Vision \\
  Istituto Italiano di Tecnologia \\
  Genova, Italy \\
  \And
  Michele Parrinello\\
  Atomistic Simulations \\
  Istituto Italiano di Tecnologia\\
  Genova, Italy \\
  \And
  Alessio Del Bue\\
  Pattern Analysis \& Computer Vision \\
  Istituto Italiano di Tecnologia \\
  Genova, Italy \\
}

\newcommand{\fix}{\marginpar{FIX}}
\newcommand{\new}{\marginpar{NEW}}

\begin{document}

\maketitle

\begin{abstract}

%
%
%


Graphs and hypergraphs provide powerful abstractions for modeling interactions among a set of entities of interest and have been attracting a growing interest in the literature thanks to many successful applications in several fields.
In particular, they are rapidly expanding in domains such as chemistry and biology, especially in the areas of drug discovery and molecule generation.
One of the areas witnessing the fasted growth is the chemical reactions field, where chemical reactions can be naturally encoded as directed hyperedges of a hypergraph.
In this paper, we address the chemical reaction classification problem by introducing the notion of a \textit{Directed Line Graph} (DLG) associated with a given directed hypergraph. On top of it, we build the \reteLONG{} (\rete{}), the first spectral-based Graph Neural Network (GNN) expressly designed to operate on a hypergraph via its DLG transformation.
The foundation of \rete{} is a novel Hermitian matrix, the \textit{\laplaciano{}}~$\mathbb{\vec L}_N$, which compactly encodes the directionality of the interactions taking place within the directed hyperedges of the hypergraph thanks to the DLG representation.
$\mathbb{\vec L}_N$ enjoys many desirable properties, including admitting an eigenvalue decomposition and being positive semidefinite, which make it well-suited for being adopted within a spectral-based GNN.
%
Through extensive experiments on chemical reaction datasets, we show that \rete{} significantly outperforms the existing approaches, achieving on a collection of real-world datasets an average relative-percentage-difference improvement of 33.01\%, with a maximum improvement of 37.71\%.

\end{abstract}

\section{Introduction}

In recent years, ground-breaking research in the graph-learning literature has been prompted by seminal works on GNNs such as~\citet{4700287, micheli2009neural, li2016gated, kipf2016semi, veličković2018graph}. 
%
%
However, representing data solely through graphs, either undirected or directed, can be limiting in many real-world applications where more complex relationships exist. In such cases, generalizations of graphs known as hypergraphs, which allow for higher-order (group) relationships among the vertices, have emerged as powerful alternatives.
Hypergraphs extend the traditional concept of a graph by allowing \textit{hyper}edges to connect an arbitrary number of nodes, thereby capturing both pairwise (dyadic) and group-wise (polyadic) interactions~\citep{schaub2021signal}.
This has naturally led to a new stream of research devoted to the investigation of Hypergraph Neural Networks (HNNs)~\citep{feng2019hypergraph, chien2021you, UniGNN, wang2022equivariant, wang2023hypergraph}.

Among many successful applications, graph and hypergraph representations have recently been applied in chemistry and biology to address various tasks such as drug discovery~\citep{bongini2021molecular}, molecule generation~\citep{hoogeboom2022equivariant}, and protein interaction modeling~\citep{jha2022prediction}.
%
Several graph-based representations have also been developed and employed for the study of chemical reactions, which has applications in areas such as reaction engineering, retrosynthetic pathway design, and reaction feasibility evaluations. 
\red{In particular, retrosynthetic modeling, where a synthetic route is designed starting from the desired product and analyzed backward, benefits greatly from accurate reaction type identification. This capability enables the elimination of unfeasible pathways, thereby streamlining the discovery of efficient routes for chemical production. This is particularly important in industries such as pharmaceutical and material sciences, where optimizing synthetic routes can lead to significant cost savings and innovation.
A similar situation holds, in reaction feasibility analysis, where predicting the likelihood of a reaction’s success based on the molecular inputs is essential for designing scalable and efficient processes.}

%
%

One of the most relevant techniques to model reactions is the directed graph~\citep{fialkowski2005architecture}, where molecules are represented as nodes and the chemical reactions are represented as directed edges from reactants to products. 
Despite its popularity, such a directed graph model suffers from some key limitations. In particular, modeling each reaction as a collection of \textit{individual} directed edges between each reactant-product pair fails to fully capture the complexity of multi-reactant or multi-product reactions, which are key to many important applications~\citep{restrepo2022chemical, garcia2023chemically}.
%
To mitigate this issue, \cite{chang2024hypergraph} proposed a hypergraph representation in which molecules are nodes and each reaction is captured by a hyperedge. However, this model lacks a mechanism to represent the directionality of a reaction, thus failing to capture the reactant/product relationship within it.
As a further attempt, \cite{restrepo2023spaces} introduced a directed hypergraph representation which is able to model both the chemical reactions structure and their directionality, where directed hyperedges model the directional interactions between reagents (heads) and products (tails), better capturing the full complexity of chemical reactions.
Let us note that this literature only focuses on modeling reaction structures without considering any form of hypergraph learning methods. We set ourselves out to developing one in this paper.

In contrast to prior studies that address node classification or link prediction tasks~\citep{dong2020hnhn, wang2023hypergraph, zhao2024dhmconv}, in this work we tackle the reaction classification problem (i.e., the problem of predicting the reaction type of a given set of reactants and products) as a {\em hyperedge classification} task.

\begin{figure}[htb!]
    \centering
    \includegraphics[width=0.77\linewidth]{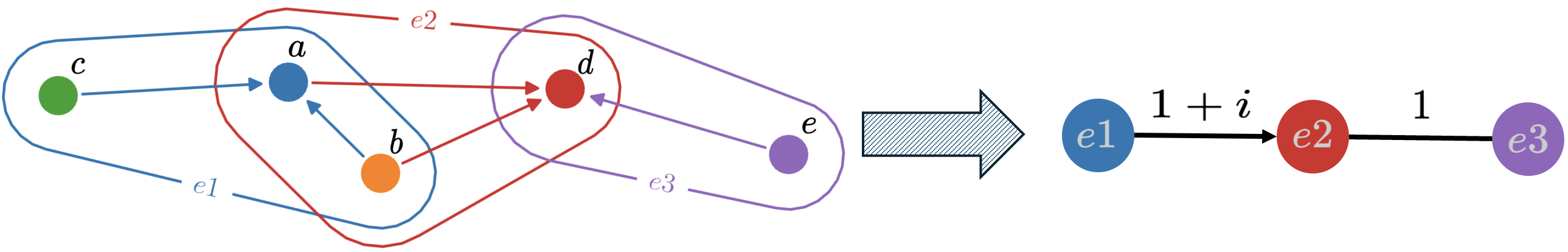} \\
    \caption{\red{Transformation from the directed hypergraph (left) to the directed line graph (right). The hyperedges of $\vec H$ become the nodes of DLG$(\vec H)$m and are connected if they overlap in $\vec H$. Complex-valued edge weights in DLG$(\vec H)$ encode $\vec H$'s directionality, as detailed in Section~\ref{sec:method}}.}
    \label{fig:linegraph}
\end{figure}

With this goal in mind, we introduce the concept of
\red{a directed line graph} of a given directed hypergraph $\vec H$: \red{the \textit{Directed Line Graph} {DLG($\vec H$)}}.
In {DLG($\vec H$)}, the vertices correspond to the hyperedges of $\vec H$, and a directed edge connects two vertices if the corresponding hyperedges in $\vec H$ share at least a vertex, as shown in Figure~\ref{fig:linegraph}. 
Since the nodes of DLG($\vec H$) correspond to hyperedges of $\vec H$, this modeling approach allows us to directly operate on hyperedge features, which are
critical for solving the reaction-classification task.
To this end, we define the \textit{Directed Line-Graph Laplacian}~$(\mathbb{\vec{L}}_N)$, a Laplacian matrix which is specifically designed to capture both directed and undirected adjacency relationships between the hyperedges of $\vec H$ via its directed line graph DLG($\vec H$). We prove that $\mathbb{\vec{L}}_N$ enjoys different key properties, among which being Hermitian (i.e., being a complex-valued matrix with a symmetric real part and a skew-symmetric imaginary one) and positive semidefinite.
These properties allows us to introduce a spectral convolutional operator for DLG($\vec H$). Thanks to the association of DLG($\vec H$) with the original directed hypergraph $\vec H$, DLG($\vec H$) serves as the foundation of \reteLONG{} (\rete{}), the first (to our knowledge) spectral-based GNN designed for the convolution of hyperedge features rather than node features.

For the task of hyperedge classification for the prediction of chemical reaction classes, transitioning from a directed hypergraphs to a directed line graph representation proves to offer significant advantages, as evidenced by our experimental results. Specifically, \rete{} achieves an average relative percentage difference improvement of 33.01\% over the second-best method across a collection of real-world datasets, with a maximum improvement of 37.71\%. We also carry out an extensive set of ablation studies, which confirm the importance of the various components of \rete{}.

\noindent \textbf{Main Contributions of This Work}
\begin{itemize}
    \item We introduce the first formal definition of
    \red{a directed line graph associated with a directed hypergraph $\vec H$: the \textit{Directed Line Graph} {DLG($\vec H$)}}.
    \item We propose the \laplaciano{}~$\mathbb{\vec{L}}_N$, a Hermitian matrix that captures both directed and undirected relationships between the hyperedges of a directed hypergraph via its DLG. We also prove that $\mathbb{\vec{L}_N}$ possesses many desirable spectral properties.
    \item We introduce \rete{}, the first spectral-based Graph Neural Network specifically designed to operate on directed line graphs associated with directed hypergraphs by directly convolving hyperedge features rather than node features.
    \item We perform an extensive experimental evaluation on the chemical reaction-classification task on real-world datasets. Our results highlight the advantages of our approach compared to other methods presented in the literature.
\end{itemize}

\section{Background}
\label{sec:bg}

An undirected hypergraph is defined as an ordered pair $H = (V, E)$, with $n := |V|$ and $m := |E|$, where $V$ is the set of vertices (or nodes) and $E \subseteq 2^{V} \setminus \{\}$ is the (nonempty) set of hyperedges.
The weights of the hyperedges are stored in the diagonal matrix $W \in \mathbb{R}^{m \times m}$, where $w_e$ is the weight of hyperedge $e \in E$ (in the unweighted case we have $W = I$).
The vertex degree $d_u$ and hyperedge degree $\delta_e$ are defined as $d_u := \sum_{e \in E: u \in e} |w_e|$ for $u \in V$, and $\delta_e := |e|$ for $e \in E$. These degrees are stored in two diagonal matrices $D_v \in \mathbb{R}^{n \times n}$ and $D_e \in \mathbb{R}^{m \times m}.$
In the case of 2-uniform hypergraphs, the matrix $A \in \mathbb{R}^{n \times n}$ is defined such that $A_{uv} = w_e$ for each $e=\{u,v\} \in E$ and $A_{uv} = 0$ otherwise; we refer to it as the {\em adjacency} matrix of the graph.
Hypergraphs where $\delta(e) = k$ for some $k \in \mathbb{N}$ for all $e \in E$ are called $k$-uniform.
Graphs are $2$-uniform hypergraphs.
%
%
Following~\cite{gallo1993directed}, we define a directed hypergraph $\vec H$ as a hypergraph where each hyperedge $e \in E$ is partitioned in
a \textit{head set} $H(e)$ and a {\em tail set} $T(e)$.
%
If $T(e)$ is empty, $e$ is an undirected hyperedge.

The relationship between vertices and hyperedges in a undirected hypergraph ${H}$ is classically represented via an incidence matrix $B$ of size $|V| \times |E|$, where 
$B$ is defined as:
\begin{equation} \label{eq:incidence}
    {B_{ve}} = 
\begin{cases}
1 & \text{if } v \in e \\
0 & \text{otherwise}
\end{cases}
\qquad v \in V, e \in E.
\end{equation}

From the incidence matrix $B$, one can derive the \textit{Signless Laplacian Matrix} $Q$
as well as its normalized version $Q_N$~\citep{chung1997spectral}:
\begin{equation}\label{eq:Q}
    Q := B W B^\top \qquad\qquad\qquad Q_N := \rsqrt{D_v} B W D_e^{-1} B^\top \rsqrt{D_v},
\end{equation}
where $W, D_e, D_v$ are the diagonal matrices defined above. 
Following~\cite{zhou2006learning}, the Laplacian for a general undirected hypergraph is defined as follows:
\begin{equation}\label{eq:delta}
    \Delta := I-Q_N.
\end{equation}

The Laplacian matrix encodes the hypergraph's connectivity and hyperedge weights.

\subsection{Graph Fourier and Graph Convolutions}

Let $\mathcal{L}$ be a generic Laplacian matrix of a given 2-uniform hypergraph $H$.
%
%
We assume that $\mathcal{L}$ admits an eigenvalue decomposition $\mathcal{L} = U \Lambda U^*$, where $U \in \mathbb{C}^{n \times n}$ represents (in its columns) the eigenvectors, $U^*$ is its conjugate transpose, and $\Lambda \in \mathbb{R}^{n \times n}$ is the diagonal matrix containing the eigenvalues. 
%
Let $x \in \mathbb{C}^n$ be a {\em graph signal}, i.e., a complex-valued function $x: V \rightarrow \mathbb{C}^n$ of the vertices of ${H}$.
We define $\hat{x} = \mathcal{F}(x) = U^* x$ as
the {\em graph Fourier transform} of $x$ and $\mathcal{F}^{-1}(\hat{x}) = U \hat{x}$ its inverse transform. 
%
The convolution $y \circledast x$ between $x$ and another graph signal $y \in \mathbb{C}^n$, acting as a \textit{filter}, in the vertex space is defined in the frequency space as $y \circledast x = U \text{diag}(U^* y) U^* x$. Letting $\hat Y := U \hat{G} U^*$ with $\hat{G} := \text{diag}(U^*y)$, we can write $y \circledast x$ as the linear operator $\hat Y x$. See~\cite{shuman2013emerging} for more details.


In the context of GNNs, explicitly learning $y$ as a {\em non-parametric filter} presents two significant limitations.
Firstly, computing the eigenvalue decomposition of $\mathcal{L}$ can be computationally too expensive~\citep{kipf2016semi}.
Secondly, explicitly learning~$y$ requires a number of parameters proportional to the input size, which becomes inefficient for high-dimensional tasks~\citep{defferrard2016convolutional}.
To address these issues, the GNN literature 
commonly employs filters where the graph Fourier transform is approximated as a degree-$K$ polynomial of $\Lambda$, with $K$ kept small for computational efficiency. For further details, we refer the reader to~\cite{kipf2016semi, defferrard2016convolutional,huang2024universal}.
%
%
This leads to a so-called \textit{localized filter}, thanks to which the output (i.e., filtered) signal at a vertex $u \in V$ is a linear combination of the input signals within $K$ edges of $u$~\citep{shuman2013emerging}.
By employing various polynomial filters and setting $K=1$ (as commonly employed in the literature), such as Chebyshev polynomials as in~\cite{hammond2011wavelets, kipf2016semi} or power monomials as used by~\cite{singh2022signed}, one obtains a parametric family of linear operators with two learnable parameters, $\theta_0$ and $\theta_1$:~\footnote{
Following w.l.o.g.~\citet{singh2022signed}, we employ the approximation $\hat G = \sum_{k=0}^K \theta_k \Lambda^k$, from which we deduce $\hat Y x = U \hat G U^* x = U (\sum_{k=0}^K \theta_k \Lambda^k) U^* x = \sum_{k=0}^K \theta_k (U \Lambda^k U^*) x = \sum_{k=0}^K \theta_k \mathcal{L}^k x$.
}
\begin{equation}\label{eq:convo}
    \hat Y := \theta_0 I + \theta_1 \mathcal{L}.
\end{equation}





\section{The Directed Line Graph and Its Laplacian}\label{sec:method}

The \textit{line graph} $L(H)$ of a generic undirected hypergraph $H$ is classically defined as the undirected graph whose vertex set is the hyperedge set of $H$. In $L(H)$, two vertices $i,j$ are adjacent---i.e., $L(H)$ contains the edge $\{i,j\}$---if and only if their corresponding hyperedges $i,j$ have a nonempty intersection~\citep{tyshkevich1998line}. 
By construction, $L(H)$ is a 2-uniform graph.
Its adjacency matrix is defined as:
\begin{equation}\label{eq:LN}
A(L(H))  := \mathbb{Q} - W D_e,
\end{equation}
where $\mathbb{Q} := B^\top B$ is, by construction, the Signless Laplacian of $L(H)$.~\footnote{This follows from the fact that the incidence matrix of $L(H)$ is $B^*$.}
%
%
%
The normalized version of $\mathbb{Q}$ and the corresponding normalized Laplacian are defined as:
\begin{equation}\label{eq:undirectedlaplacian}
    \mathbb{Q} := \sqrt{W}B^\top B \sqrt{W} \qquad\qquad {\mathbb{Q}_{N}} := \rsqrt{{D}_e} \sqrt{W} {B}^\top {D}_v^{-1}{B} \sqrt{W} \rsqrt{{D}_e}
    \qquad\qquad
    \mathbb{L}_N := I - \mathbb{Q}_{N}.
\end{equation}
Notice that, from~\eqref{eq:Q}, one can define the weighted version of $B$ as $B \sqrt{W}$. The definitions in~\eqref{eq:undirectedlaplacian} rely on the same matrix, but transposed.

To the best of our knowledge, the literature does not offer any formal definition for the line graph associated with a (weighted) directed hypergraph $\vec{H}$ \red{(it does only for the undirected case)}.
The availability of such a definition could be crucial for tasks where the hyperedge direction is important.

To address this gap, we first define a
complex-valued incidence matrix $\vec{B}$ which preserves the inherent directionality of $\vec{H}$:
\begin{equation} \label{eq:myproposal}
    \vec{B}_{ve} := 
\begin{cases}
    1 & \text{if } v \in H(e), \\
    -\ii & \text{if } v \in T(e), \\
    0 & \text{otherwise}.
\end{cases}
    \qquad v \in V, e \in E.
\end{equation}

Building on $\vec{B}$, we propose the following definition for the directed line graph associated with a directed hypergraph $\vec H$:
\begin{definition}\label{def:DLG}
    The Directed Line Graph $DLG(\vec{H})$ of a directed hypergraph $\vec{H}$ is a 2-uniform hypergraph whose vertex set corresponds to the hyperedge set of $\vec{H}$
    and whose adjacency matrix is the following complex-valued skew-symmetric matrix:
\begin{equation}\label{eq:adj}
A(DLG(\vec{H})) = 
\sqrt{W}\vec{B}^* \vec{B}\sqrt{W} - W D_e.
\end{equation}
\end{definition}





Using \eqref{eq:adj} of definition~\ref{def:DLG} and equations~\ref{eq:LN}--\ref{eq:undirectedlaplacian}, we obtain the following formulas for the normalized Signless Laplacian $\mathbb{\vec{Q}}_{N}$ and the normalized Laplacian $\mathbb{\vec{L}}_N$ of $DLG$, which we refer to by
{\em Signless Directed Line-Graph} and {\em \laplaciano{}}:
%
\begin{equation}\label{eq:myproposal2}
    \mathbb{\vec{Q}}_{N} :=  \rsqrt{\vec{D}_e} \sqrt{W} \vec{B}^* \vec{D}_v^{-1} \vec{B} \sqrt{W} \rsqrt{\vec{D}_e} \qquad \mathbb{\vec{L}}_N :=  I - \mathbb{\vec{Q}}_{N}.
\end{equation}




To better understand how $\mathbb{\vec{L}}_N$ encodes the directionality of $\vec H$, we illustrate its definition in scalar form for a pair of hyperedges $i, j \in E$ (which correspond to vertices in $DLG(\vec H)$):
%
\begin{eqnarray}\label{eq:laplacian-expanded}
    \mathbb{\vec L}_N(ij)= 
    \left\{
    \begin{array}{lr}
    \displaystyle
    1 - \sum_{
    u \in i}\frac{w_i}{d_u\delta_i}& i = j\\
    \displaystyle
    \left(-\hspace{-.4cm}\sum_{\substack{
    u \in H(i) \cap H(j) \\ \vee u \in T(i) \cap T(j)}} \hspace{-.5cm}\frac{\sqrt{w_i} \sqrt{w_j}}{d_u} - 
    \ii 
    \left(\sum_{\substack{
    u \in H(i) \cap T(j)}}
    \hspace{-.3cm}\frac{\sqrt{w_i} \sqrt{w_j}}{d_u} -
    \sum_{\substack{
    u \in T(i) \cap H(j)}} 
    \hspace{-.3cm}\frac{\sqrt{w_i} \sqrt{w_j}}{d_u}\right)\right) \frac{1}{\sqrt{\delta_i}} \frac{1}{\sqrt{\delta_j}} & i \neq j
    \end{array}
    \right.
\end{eqnarray}
%
\red{When $i = j$, \red{we are in the self-loop part of the equation and} $\mathbb{\vec L}_N(ij)$} weights \red{hyperedge $i$ proportionally to its weight $w_i$ and inversely proportionally to its density and the density of its nodes}.
%
\red{When $i \neq j$, $\mathbb{\vec L}_N(ij)$'s value} depends on the interactions between the hyperedges of $\vec H$ (which correspond to the nodes of $DGL(\vec H)$).
Let $u \in V$ be a node and $i, j \in E$ be two hyperedges in the hypergraph $\vec H$.
If $u$ belongs to the head set of both the hyperedges (i.e., $u \in H(i) \cap H(j)$) or to the tail set of both (i.e., $u \in T(i) \cap T(j)$), its contribution to the real part of $\mathbb{L}_N(ij)$, $\real(\mathbb{\vec{L}}_N(ij))$, is negative. 
For the undirected line graph associated with an undirected hypergraph, this is the only contribution, consistent with the behavior of $\mathbb{L}_N$ (as described in \eqref{eq:undirectedlaplacian}).
%
If $u$ takes opposite roles in hyperedges $i$ and $j$, i.e, it belongs to the head set in $i$ and to the tail set in $j$ or {\em vice versa}, it contributes to the imaginary part of $\mathbb{L}_N$, $\imaginary(\mathbb{\vec{L}}_N(ij)$, negatively when $u \in H(i) \cap T(j)$, and positively when $u \in T(i) \cap  H(j)$.
Consequently $\imaginary(\mathbb{\vec L}_N (ij))$ coincides with the {\em net} contribution of all the vertices that are shared between the hyperedges $i$ and $j$.
An example illustrating the construction of $\mathbb{\vec{L}}_N$ for a directed line graph associated with a directed hypergraph is provided in Appendix~\ref{appx:complex_laplacian}.
Let us point out that the behavior of {\em \laplaciano{}} differs from every (to the best of our knowledge) Laplacian matrix previously proposed in literature (see Appendix~\ref{appx:theorem} for more details).

With the following theorem, we show that $\mathbb{\vec L}_N$ is a generalization of $\mathbb{L}_N$ (defined in~\eqref{eq:undirectedlaplacian}) from the undirected to the directed case:
\Copy{thm:generalizeHyper}{
\begin{theorem}\label{thm:generalizeHyper}
    If $\vec{H}$ is undirected (i.e., $\vec{H} = H$), $\mathbb{\vec{L}}_N =  \mathbb{L}_N$ and $\mathbb{\vec Q}_{N} = \mathbb{Q}_{N}$ holds.
\end{theorem}
}


%
The {\em \laplaciano{}} enjoys several properties. 
%
%
%
First, to be able to adopt our Laplacian within a convolution operator in line with~\cite{kipf2016semi} and other literature approaches~\citep{zhang2021magnet,fiorini2023sigmanet}, we must show that our Laplacian is positive semidefinite. For this, we work out the expression for the $2$-Dirichlet energy function associated with it. Such a function coincides with the Euclidean norm $||x||^2_{\mathbb{\vec L}_N}$ induced by $\mathbb{\vec L}_N$ for a signal $x \in \mathbb{C}^n$:
\Copy{thm:dirichlet}{
    \begin{theorem}\label{thm:dirichlet}
    Letting $\mathbf{1}$ be the indicator function, the Euclidean norm induced by $\mathbb{\vec L}_N$ of a complex-valued signal $x = a + \ii b \in \mathbb{C}^{n}$ reads:
    \begin{align}\label{eq:all-together}
     %
     \nonumber
     \frac{1}{2} \sum_{u \in V}  \frac{1}{d(u)} \sum_{i, j \in E} \sqrt{w(i)} \Bigg(
     & \left(\left(\frac{a_i}{\sqrt{\delta(i)}} - \frac{a_j}{\sqrt{\delta(j)}} \right)^2 + \left(\frac{b_i}{\sqrt{\delta(i)}} - \frac{b_j}{\sqrt{\delta(j)}}\right)^2\right) \mathbf{1}_{ 
     u \in H(i) \cap H(j) \vee u \in T(i) \cap T(j)}\\
     %
     \nonumber
     & + \left(\left(\frac{a_i}{\sqrt{\delta(i)}} - \frac{b_j}{\sqrt{\delta(j)}} \right)^2 + \left(\frac{a_j}{\sqrt{\delta(j)}} + \frac{b_i}{\sqrt{\delta(i)}}\right)^2\right) \mathbf{1}_{u\in H(i) \cap T(j)} \\
     & + \left( \left(\frac{a_i}{\sqrt{\delta(i)}} + \frac{b_j}{\sqrt{\delta(j)}} \right)^2 + \left(\frac{a_j}{\sqrt{\delta(j)}} - \frac{b_i}{\sqrt{\delta(i)}}\right)^2\right) \mathbf{1}_{u\in T(i) \cap H(j)} 
     \Bigg) \sqrt{w(j)}.
    \end{align}
    \end{theorem}
}

Since the function in Theorem~\ref{thm:dirichlet} is a real-valued sum of squares, we deduce the following spectral property for $\mathbb{\vec L}_N$:
%
\Copy{thm:psdL}{
    \begin{corollary}\label{thm:psdL} 
    $\mathbb{\vec L}_N$ is positive semidefinite.
    \end{corollary}
}


From equation~\ref{eq:myproposal2}, we have that $\mathbb{\vec L}_N = I - \mathbb{\vec Q}_N$. Thanks to Theorem~\ref{thm:psd}, we show next that $\mathbb{\vec Q}_{N}$ has a nonnegative spectrum:
%
\Copy{thm:psd}{
    \begin{theorem}\label{thm:psd}
    $\mathbb{\vec Q}_{N}$ is positive semidefinite.
    \end{theorem}
}

By applying Theorem~\ref{thm:psd} and Corollary~\ref{thm:psdL}, we can derive upper bounds on the spectra of $\mathbb{\vec L}_N$ and $\mathbb{\vec Q}_{N}$:
%
%
\Copy{thm:eigenvalues}{
    \begin{corollary}\label{thm:eigenvalues}
    $\lambda_{\max}(\mathbb{\vec L}_N) \leq 1$ and $\lambda_{\max}(\mathbb{\vec Q}_{N}) \leq 1$.
    \end{corollary}
}


The proofs of the theorems and corollaries of this section can be found in Appendix~\ref{appx:theorem}.

\section{The \reteLONG~(\rete)}\label{sec:rete}

The properties of the proposed Laplacian make it possible to derive a well-defined spectral convolution operator from it. In this work, this operator is integrated into the \reteLONG{} (\rete{}). 
%
Specifically, based on~\eqref{eq:convo}, by setting $\mathcal{L} = \mathbb{\vec L}_N$, the convolution operator is defined as $\hat{Y} x = \theta_0 I + \theta_1 \mathbb{\vec L}_N$.
%
The advantage of adopting two parameters $\theta_0, \theta_1$ within \rete{}'s localized filter is explained by the following result:
\Copy{thm:parameters}{
    \begin{proposition}\label{thm:parameters}
    The convolution operator derived from \eqref{eq:convo} by setting $\mathcal{L} = \mathbb{\vec{L}}_N$ with parameters $\theta_0$ and $\theta_1$
    is the same as the convolution operator obtained by using $\mathcal{L} = \mathbb{\vec{Q}}_N$ with parameters are rewritten as $\theta_0' = \theta_0 + \theta_1$ and $\theta_1' = -\theta_1$.
    \end{proposition}
}
%
This shows that \rete{}, by selecting appropriate values for $\theta_0$ and $\theta_1$, can leverage either $\mathbb{\vec{L}}_N$ or $\mathbb{\vec{Q}}_N$ as convolution operator to maximize the performance on the task at hand.

We define $X \in \mathbb{C}^{m \times c_0}$ as a $c_0$-dimensional graph signal (a graph signal with $c_0$ input channels), which we compactly represent as a matrix. 
This matrix serves as the feature matrix of the hyperedges of $\vec H$ which we construct from the feature matrix of the nodes $X' \in \mathbb{C}^{n \times c_0}$ of $\vec H$. Specifically, inspired by the operation used in the \textit{reduction component} for graph pooling~\citep{grattarola2022understanding}, we define the feature matrix for the vertices of $DGL(\vec H)$ as $X = \vec{B}^* X'$. This approach combines features through summation, based on the topology defined by $\vec{B}$. See Appendix~\ref{appx:features} for more details.

In our network, the scalar parameters $\theta_0$ and $\theta_1$ are subsumed by 
two operators $\Theta_0, \Theta_1 \in \mathbb{C}^{c_0 \times c}$ which we use to carry out a linear transformation on the feature matrix $X$. A similar transformation, which can also increase or decrease the number of channels of $X$, is adopted in other GNNs such as MagNet~\citep{zhang2021magnet}.
%
\rete{} features $\ell$ convolutional layers. The output $Z \in \mathbb{C}^{m \times c'}$ of any such layer adheres to the following equation:
\begin{equation}\label{eq:convolution}
    Z (X) = \phi\left(IX\Theta_0 + \mathbb{\vec L}_N X\Theta_1\right),
\end{equation}
where $\phi$ is the activation function. Following~\citep{fiorini2023sigmanet,fiorini2024graph}, \rete{} employs a complex \textit{ReLU} where $\phi(z) = z$ if $ \real(z) \geq 0$ and $\phi(z) = 0$ otherwise, with $z \in \mathbb{C}$.
\rete{} also utilizes a residual connection for every convolutional layer except the first one, a choice which helps prevent oversmoothing and has been proven to be helpful in a number of works, including~\citep{he2016deep, kipf2016semi}.
After the convolutional layers, following~\cite{zhang2021magnet}, we apply an {\em unwind} operation where we transform $Z(X) \in \mathbb{C}^{m \times c'}$ into $(\real(Z (X)) || \imaginary(Z (X))) \in \mathbb{R}^{n \times 2c'}$, where $||$ is the concatenation operator.
To obtain the final results, \rete{} features $S$ linear layers, with the last one employing a Softmax activation function.

\paragraph{Complexity of \rete.} Let us assume (w.l.o.g.) that each of \rete{}'s convolutional layers has $c$ input and output channels, while the last layer has $c$ input and $c'$ output channels ($c'$ is also the number of input channel to the linear layers). Let $d$ be number of output channel of the last linear layer (where $d$ is the number of classes to be predicted).
With $\ell$ convolutional layers and $S$ linear layers, \rete{}'s complexity is
$
O(m n c_0) 
+ O(\ell (m^2 c + m c^2) + mc  + (S-1)(m c'^2) + m c' d + md)$.
Assuming $O(c) = O(c') = O(d) = \bar c$, we have a complexity of $O(\ell (m^2 \bar c) + (\ell + S) (m \bar c^2))$.
%
%
This shows that \rete{} has a quadratic complexity w.r.t. the number of hyperpedges $m$ and the asymptotic number of channels $\bar c$.
For more details, see Appendix~\ref{appx:complexity}.






\section{Experimental results}

\red{We present three real-world datasets, the baseline models, and the results on the chemical reaction classification task, where we predict the reaction type based on a given set of molecules.}



%

%


\subsection{Datasets}

\begin{figure}[h!]
    \centering
    \includegraphics[width=0.9\linewidth]{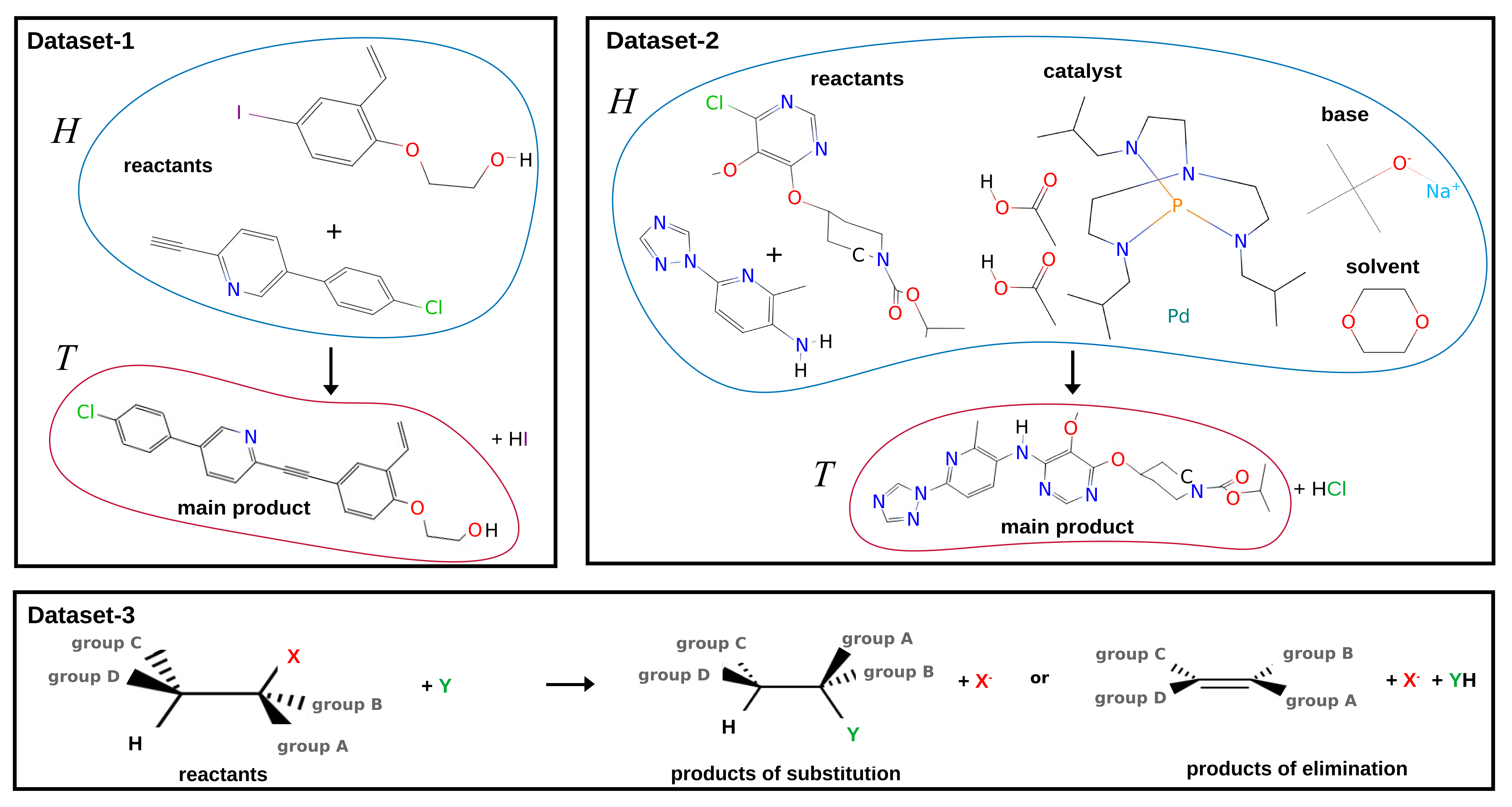} \\
    \caption{
    {\bf (Upper panel, left)}:
    \red{example} from {\tt Dataset-1}. C--C bond formation via reaction of alkyne with alkyl halide; only bi-molecular reactant and main product are taken into account (any byproduct is omitted).
    {\bf (Upper panel, right)}: example from {\tt Dataset-2}. C--N bond formation via Buchwald-Hartwig amination; 
    apart from bi-molecular reactant (amine and aryl halide) and main product, catalyst (palladium compound), solvent (dioxane) and base (sodium tert-butoxide) structures are also present.
    Chemical elements: carbon (C), nitrogen (N), oxygen (O), hydrogen (H), chlorine (Cl), iodine (I), sodium (Na), phosphorus (P) and palladium (Pd). Single, double and triple black lines: bonds between C atoms.
    \textit{H}, \textit{T}: Head and Tail of the directed hypergraph.
    {\bf (Lower panel)}: schematic representation of {\tt Dataset-3} elements. Left side: reactants; right side: competitive outcomes between bimolecular nucleophilic substitution (S$\mathrm{_N}$2) or bimolecular elimination (E2). Thus, each element is composed either of a bi-molecular reactant and a bi-molecular product (S$\mathrm{_N}$2 class), or a bi-molecular reactant and a tri-molecular product (E2 class). X and Y: leaving group and nucleophile agent. Groups A-D: different substituents attached to the alkane carbon backbone (black).
   }
    \label{fig:datasets}
\end{figure}

We test \rete{} on the most common organic chemistry reaction classes, namely a variety of chemical transformations that are fundamental to both research and industrial chemistry.
Those include molecular rearrangements, such as the interconversion (substitution) or the elimination of molecular substituents, as well as the introduction of specific functional groups (e.g., acyl, alkyl, or aryl groups) in a chemical compound. Other important reactions classes involve the  formation of certain bond-types (e.g., carbon-carbon: C--C) or structures (e.g., heterocyclic compounds), the change in the oxidation state of a molecular species (oxidation-reduction processes), and the protection/deprotection of functional groups, allowing to temporarily block a specific reactive site at a certain step in a synthetic route. 
For our study, we rely on a standard dataset ({\tt Dataset-1}) and additionally construct two new ones ({\tt Dataset-2} and {\tt Dataset-3}):---see Figure~\ref{fig:datasets}.

\noindent \textbf{Dataset-1.} As main source of data, we use the reactions from USPTO granted patents
~\citep{lugo2021classification}, which is the most widely used dataset for retrosynthesis problems and contains about 480K reactions.     
After removing duplicates and erroneous reactions, we select a subset, namely {\tt Dataset-1}, comprising 50K atom-mapped reactions belonging to 10 different classes. An example component from {\tt Dataset-1} is reported in Figure~\ref{fig:datasets}, left upper panel. The composition of the dataset is detailed in Table~\ref{tab:uspto}, Appendix~\ref{appx:dataset}.

\noindent \textbf{Dataset-2.} This dataset is the result of the merging of data from five different sources and contains 5300 reactions. It presents a smaller number of reaction types, but a larger variety of substituents and reaction conditions, such as the presence of solvent or catalyst, hence providing additional complexity on some specific classes for the model to predict. 
Figure~\ref{fig:datasets}, upper right panel illustrates an example from it.
Given that some elements are shared across the data sources, we combine them into three major classes. 
The elements of the first class (C--C bond formation) 
are extracted from two separated collections present in the Open Reaction Database (ORD) Project~\citep{kearnes2021open}. Those are the~\cite{reizman2016suzuki} data  for the Pd-catalyzed Suzuki–Miyaura cross-coupling reactions and a vast collection of Pd-catalyzed imidazole-aryl coupling reactions, via C-H arylation. 
The elements of \textit{Class 2} (N-arylation) includes data of Pd-catalyzed N-arylation (Buchwald-Hartwig) reactions from the AstraZeneca ELN dataset, also generated from the ORD website. This class has been further divided in 3 sub-classes according to the nature of the aryl halide used for the coupling.   
Finally, the third class contains an ORD collection of data for amide bond formation processes. 
Details about {\tt Dataset-2} composition are reported in  Table~\ref{tab:ord}, Appendix~\ref{appx:dataset}. 

\noindent \textbf{Dataset-3.} Since the wo datasets listed so far only include single-product reactions, in order to test the model on a highly complex task we add a third collection, {\tt Dataset-3}, comprised of double-product bimolecular nucleophilic substitution (S$\mathrm{_N}$2) and triple-product bimolecular elimination (E2) reaction classes, extracted from~\cite{von2020thousands} and totaling 649 competitive reactions. 
A schematic representation of {\tt Dataset-3} elements is reported in Figure \ref{fig:datasets}, lower panel. Further details can be found in Appendix~\ref{appx:dataset}.

In all three datasets, the node features are build based on \textit{Morgan Fingerprints} (MFs) \cite{Rogers2010ECFP}, which are one of the most widely used molecular descriptors. MFs encode a molecule by capturing the presence or absence of specific substructures (fragments) within the molecular graph. The algorithm iteratively updates the representation of each atom based on its local environment, enclosed within a radius.
A radius of $r$ indicates that the environment up to $r$ bonds away from each atom is incorporated into the final representation.


\subsection{Baselines and Experimental Details}

We evaluate the performance of \rete{} against 12 state-of-the-art (baseline) methods: \textit{i)} \red{Undirected Hypergraph Neural Networks (HNNs)}: HGNN~\citep{feng2019hypergraph}, HCHA\footnote{Among the many versions of HCHA in~\cite{dong2020hnhn}, we use the one implemented in~\url{https://github.com/Graph-COM/ED-HNN}, which coincides with HGNN$^+$~\citep{gao2022hgnn}.}~\citep{bai2021hypergraph}, HCHA with the attention mechanism~\citep{bai2021hypergraph}, HNHN~\citep{dong2020hnhn}, 
UniGCNII~\citep{UniGNN}, HyperDN~\citep{tudisco2021nonlinea}, AllDeepSets~\citep{chien2021you}, AllSetTransformer~\citep{chien2021you}, LEGCN~\cite{yang2022hypergraph}, ED-HNN~\citep{wang2022equivariant}, and PhenomNN~\citep{wang2023hypergraph}; \textit{ii)} \red{Directed HNN}: DHM~\citep{zhao2024dhmconv}. 
Since all the competitors operate directly on the undirected or directed hypergraph, we apply the feature transfer operation $X = \vec{B}^* X'$ described in Section~\ref{sec:method} (more details in Appendix~\ref{appx:features}) after the convolutional layers. After this step, each method is equipped with $\ell$ linear layers. 
The hyperparameters of these baselines and of our proposed model are selected via grid search (see Appendix~\ref{appx:experiment}).
%
%
The datasets are split into 50\% for training, 25\% for validation, and 25\% for testing.
The experiments are conducted with 5-fold cross-validation, reporting the average F1-score across the splits. We choose the F1-score as evaluation metric due to the class imbalance naturally present in the datasets.
Throughout the tables contained in this section, the best results are reported in \textbf{boldface} and the second best are \ul{underlined}. 

\subsection{Results}


\paragraph{Quantitative.}

The F1-score along with the relative standard deviation across different methods, datasets, and folds is presented in Table~\ref{tab:result}. The results show that, across the three datasets, \rete{} achieves an average additive performance improvement over the best-performing competitor 
of approximately 23.51 percentage points. In terms of Relative Percentage Difference (RPD)\footnote{The RPD of two values $P_1, P_2$ is the percentage ratio of their difference to their average, i.e., $|P_1-P_2| / \frac{P_1+P_2}{2}\%$.}, we have an average RPD improvement of 33.01\%.
\rete{} achieves the best improvement on {\tt Dataset-3}, with an average RPD improvement of approximately 37.71\% and an average additive improvement of 31.65 percentage points.
A  clear trend emerges: HNNs-based methods designed for undirected hypergraphs consistently underperform compared to DHM, which is the only method specifically designed for handling directed hypergraphs. Crucially, our proposed \rete{}, which operates on the directed line graph, surpasses all the competitors in performance, \red{incuding DHM}.

\begin{table}[htb!]
\centering
\caption{Mean F1-score and standard deviation obtained on the hyperedge classification task.}
\resizebox{\columnwidth}{!}{%
\begin{tabular}{clrrr}
\hline
Topology & Method & {\tt Dataset-1} & {\tt Dataset-2} & {\tt Dataset-3}\\ \hline
&HGNN & 9.71 $\pm$ 3.02 &  36.40 $\pm$ 7.27 & 64.97 $\pm$ 1.36\\
&HCHA/HGNN$^+$ & 9.99 $\pm$ 1.91 & 39.89 $\pm$ 4.93 & 63.46 $\pm$ 2.58\\
&HCHA w/ Attention & 9.90 $\pm$ 2.25 & 11.32 $\pm$ 0.16  & 35.55 $\pm$ 1.31\\
\multirow{5}{*}{\rotatebox{0} {Hypergraph}}&HNHN & 6.95 $\pm$ 0.97 & 25.04 $\pm$ 3.45  & 52.97 $\pm$ 5.17\\
&UniGCNII & 8.20 $\pm$ 2.39 & 29.86 $\pm$ 0.31 & 50.97 $\pm$ 6.84\\
&HyperND & 4.63 $\pm$ 0.04 & 28.98 $\pm$ 0.46 & 52.71 $\pm$ 12.32\\
&AllDeepSets& 7.64 $\pm$ 2.23 & 30.45 $\pm$ 0.27  & 51.72 $\pm$ 5.99 \\
&AllSetTransformer & 8.63 $\pm$ 2.62 & 30.67 $\pm$ 0.57 & 49.24 $\pm$ 3.98\\
&ED-HNN & 9.19 $\pm$ 1.43 & 30.47 $\pm$ 0.56 & 50.52 $\pm$ 10.17\\
&PhenomNN & 8.33 $\pm$ 2.77 & 29.43 $\pm$ 0.39 & 51.82 $\pm$ 9.33 \\ \hline 
Directed Hypergraph &DHM & \ul{46.04 $\pm$ 0.58} & \ul{59.31 $\pm$ 4.04} & \ul{68.10 $\pm$ 3.60}\\ \hline
Directed-Line Graph & \textbf{\rete{}} & \textbf{60.55 $\pm$ 0.80} & \textbf{83.67 $\pm$ 3.41} & \textbf{99.75 $\pm$ 0.34}\\
\hline
\end{tabular}%
}
\label{tab:result}
\end{table}

\paragraph{Qualitative.} 

\begin{figure}[htb!]
    \centering
    \includegraphics[width=0.9\linewidth]{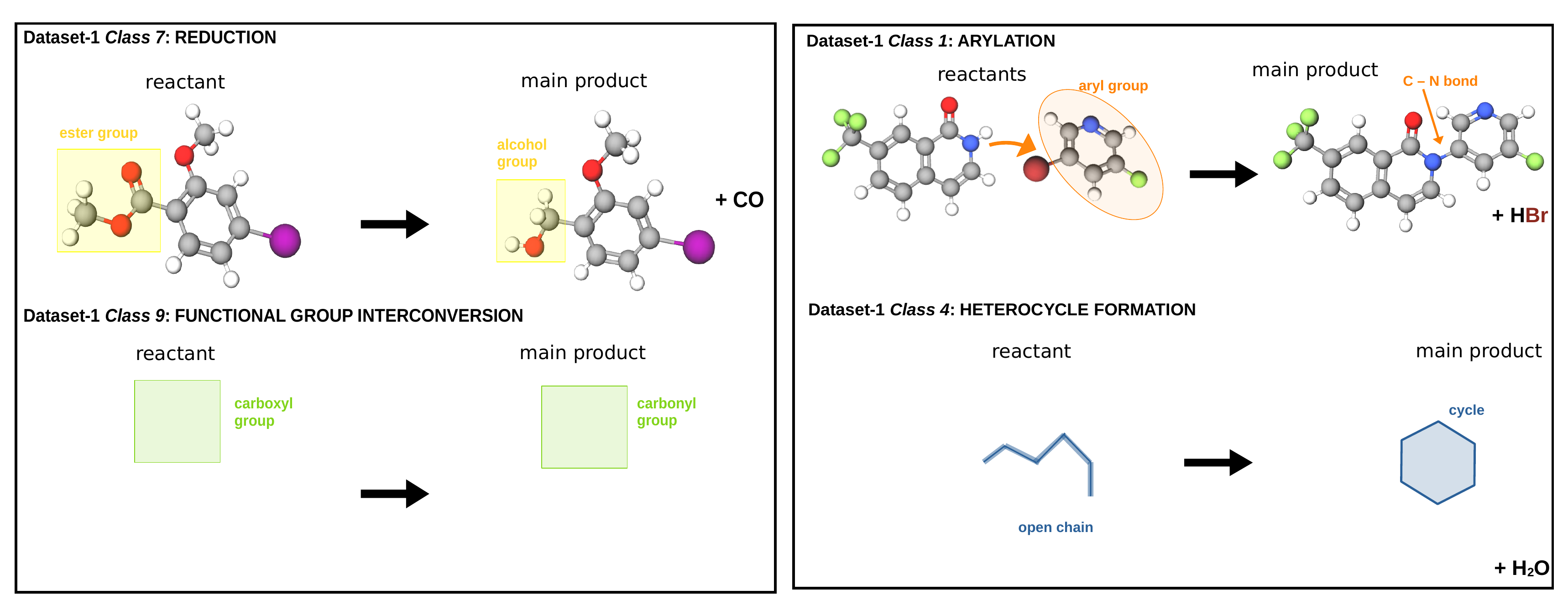} \\
    \caption{
    Ball-and-stick 3D model of {\tt Dataset-1} mislabeled pairs of reaction classes.
    Color code: grey for carbon, red for oxygen, blue for nitrogen, purple for iodine, green for chlorine, light green for fluorine, brown for bromide, and white for hydrogen.
    {\bf (Left panel, upper)}: Reduction from a ester to a alcohol substituent on a 6-carbon atoms ring.
    {\bf (Left panel, lower)}: Functional group interconversion from carboxyl to carbonyl group in the analog hexagonal structure. 
    {\bf (Right panel, upper)}: arylation reaction between a amine compound and a aryl halide, yielding a C--N bond in the final product.
    {\bf (Right panel, lower)}: heterocycle formation via amide intramolecular condensation, producing a hexagonal ring containing a heteroatom (nitrogen). }
    \label{fig:2}
\end{figure}

To gain deeper insights into the capability of \rete{} of classifying different reaction types, we analyze the confusion matrices for {\tt Dataset-1} and {\tt Dataset-2}. The results of this analysis are presented in Figure~\ref{fig:dataset-1} and Figure~\ref{fig:dataset-2} in Appendix~\ref{appx:confusionmatrix}. 
The confusion matrix for {\tt Dataset-1} reveals
\red{that, while most of the classes are predicted extremely well}, e.g., Protection and Functional group addition reactions 
\red{(accuracy} of 88\% and 77\%, respectively),
\red{some are predicted not as well}, e.g., Functional group interconversion (41\%). 
To better understand this behavior, we conducted a thorough inspection of the structural features of {\tt Dataset-1}'s components, selecting several elements from pairs of classes among which the model yields the highest uncertainty. Two example cases are reported in Figure~\ref{fig:2}.
\red{Overall, our analysis reveals that the pair of classes which are subject to the higher degree of confusion are, structurally, highly similar, which well explains the poorer performance that \rete{} achieves on them, as we illustrate in the following.}
The left panel illustrates the mislabeling of \textit{Class 9} (Functional group interconversions, correctly predicted in 41\% of the cases) with \textit{Class 7} (Reductions, incorrectly predicted in 14\% of the cases), while the right panel presents an example of \textit{Class 4} (Heterocycle formations, correctly predicted in 44\% of the cases) with \textit{Class 1} (Arylations, incorrectly predicted 30\% of the cases).    
Notably, in these examples, both the main backbone structure of the molecules and the substituent groups (the segments affected by the reactive process, highlighted in the figure) exhibit a high degree of similarity between the two classes.
In the left panel, the reactants of both classes present a 6-carbon ring (in grey) as well as a iodine substituent (in purple).
The atoms composing the highlighted groups are also of the same types.        
On the other hand, in the right panel, the majority of the constituent parts of the products are in common between the two classes. Specifically, despite the outcome of \textit{Class 4} is the formation of a heterocycle, i.e., a hexagonal ring containing a heteroatom (nitrogen, in blue), such a geometrical feature is also present in \textit{Class 1} arylation product, as the resulting molecule presents two heterocycles rings.
Similar considerations apply to the incorrect labeling of {\tt Dataset-2} N-arylation sub classes, where the main difference between the reactants lies in the nature of the aryl halide that participates in the coupling reaction.
\red{In summary, we conclude that the model demonstrates strong predictive performance across the majority of the classes, although a few, particularly those with shared elements, remain challenging to differentiate. Nevertheless, we are confident that \rete{} will prove highly valuable to the chemistry community, allowing for the categorization of existing data sources as well as for planning new synthetic routes.}


\paragraph{Ablation study.}

Table~\ref{tab:ablation} presents the results of an ablation study carried out on \rete{} to assess the importance of directionality in \rete{}'s line graph. To do this, we test \rete{} using an undirected line graph and demonstrate that \rete{} consistently outperforms its undirected counterpart on all three data sets. This indicates that directionality plays a crucial role in solving the chemical reaction classification task.
Focusing on~\eqref{eq:convolution}, we test \rete{} under two conditions: \textit{i)} using $\mathbb{\vec Q}_N$ instead of $\mathbb{\vec L}_N$, and \textit{ii)} setting $\Theta_0 = 0$, thus nullifying the first term in~\eqref{eq:convolution}. The first comparison shows identical results across all datasets, thus providing a computational confirmation of the results of Proposition~\ref{thm:parameters}, while the restricted version of \rete{} with $\Theta_0 = 0$ performs worse.
Finally, we assess the architectural choice related to the incorporation of skip connections. While \rete{} without skip connections exhibits a slight drop in performance, the results remain close to those of the original architecture.

\begin{table}[htb!]
\caption{\red{Ablation study. Average F1-score and standard deviation are reported.}}
\centering
\resizebox{0.8\columnwidth}{!}{%
\begin{tabular}{Hlrrr}
\hline
Topology & Method &{\tt Dataset-1} & {\tt Dataset-2} & {\tt Dataset-3}\\ \hline
& \textbf{\rete{}} & \textbf{60.55 $\pm$ 0.80} & \textbf{83.67 $\pm$ 3.41} & \textbf{99.75 $\pm$ 0.34}\\
Line Graph & \rete{} w/o directionality & 52.07 $\pm$ 1.61 & 70.19 $\pm$ 0.65 & {81.65 $\pm$ 8.39}\\
Line Graph & \rete{} w/ Signless Laplacian & \ul{60.24 $\pm$ 0.36} & \ul{82.86 $\pm$ 1.96}  & \textbf{99.75 $\pm$ 0.55} \\
Line Graph & \rete{}  w/ $\Theta_0 = 0$ & 53.82 $\pm$ 0.74 & 75.68 $\pm$ 3.59  & {91.45 $\pm$ 2.36}\\
Line Graph & \rete{}  w/o skip-connection & 56.38 $\pm$ 3.02 & 80.63 $\pm$ 3.54  & \ul{99.63 $\pm$ 0.34}\\
\hline
\end{tabular}%
}
\label{tab:ablation}
\end{table}

\section{Conclusions}

We introduced the \reteLONG{} (\rete{}), the first spectral GNN specifically designed
to operate on directed line graphs associated with directed hypergraphs by directly convolving hyperedge features.
\rete{} leverages a novel complex-valued Laplacian matrix, the \textit{\laplaciano{}}, which is a Hermitian matrix encoding the interactions among the hyperedges of a hypergraph using complex numbers. This formulation allows for the natural representation of both directed and undirected relationships between the hyperedges, capturing rich structural information.
Our proposed \rete{} network utilizes this new Laplacian matrix to perform spectral convolutions on the line graph featuring both undirected and directed edges.
Via the Directed Line Graph representation, our proposed model enables the seamless integration of the directionality present in the hypergraph at hand, which is crucial for accurately modeling various real-world phenomena involving asymmetric high-order interactions.

We evaluated our approach on the chemical reaction classification problem using three real-world datasets. In these experiments, we demonstrated the superiority of \rete{}, which achieved an average relative percentage difference improvement of 33.01\% over the second-best method across the three datasets.
This highlights the importance of directly convolving the hypergraph features on the directed line graph, instead of doing so in the undirecetd/directed hypergraph. Though an ablation study, we demonstrated the relevance of encoding directional information via the directed line graph associated with a directed hypergraph as opposed to ignoring it. We also provided a qualitative analysis \rete{}'s results in light of the underlying chemical reaction classification task.

In light of the promising results we obtained and as a future perspective, 
\red{we would like to address more complex and challenging tasks, such as retrosynthetic planning and reaction discovery, which require sophisticated analysis and deeper insights into the underlying chemical processes.}

\newpage
\bibliographystyle{plainnat}
\bibliography{BIB}

\newpage 
\appendix

%
%
%

\section{Properties of Our Proposed Laplacian}\label{appx:theorem}

This section contains the proofs of the theorems and corollaries reported in the main paper.

\setcounter{theorem}{0}
\setcounter{proposition}{0}
\setcounter{corollary}{0}

\Paste{thm:generalizeHyper}
\begin{proof}
    Since ${H} = (V, E)$ is an undirected hypergraph, $\vec B$ is binary and only takes values 0 and 1 (rather than being ternary and taking values $0, 1, -\ii$), defining an undirected line graph $L(H)$. In particular, for each edge $e \in E$ we have $\vec{B}_{ue}=1$ if either $u \in H(e)$ or $u \in T(e)$ and $\vec B_{ue} = 0$ otherwise. Consequently, the directed incident matrix $\vec{B}$ is identical to the non-directed incidence matrix $B$, i.e., $\vec{B} = B$. Thus, by construction, $\mathbb{\vec L}_N =  \mathbb{L}_N$ and $\mathbb{\vec Q}_N =  \mathbb{Q}_N$.
\end{proof}


\Paste{thm:dirichlet}
\begin{proof}
\small
 \begin{align*}
&& x^*\mathbb{\vec L}
_Nx = & \sum_{i \in E} x_i^* x_i - \sum_{i, j \in E}   \sum_{u \in V}   \frac{1}{d(u)} \frac{\rsqrt{w(i)}\vec B(u, i)^* \vec B(u, j) \rsqrt{w(j)}}{\sqrt{\delta(i)} \sqrt{\delta(j)}} x_i x_j^*    \\ 
&& = & \sum_{i \in E} x_i^* x_i -   \sum_{u \in V} \sum_{i, j \in E} \frac{1}{d(u)} \frac{\rsqrt{w(i)}\vec B(u, i)^* \vec B(u, j) \rsqrt{w(j)}}{\sqrt{\delta(i)} \sqrt{\delta(j)}} x_i x_j^*  \\    
&& = & \sum_{i \in E} x_i^* x_i -  \sum_{u \in V}  \frac{1}{d(u)} \sum_{i, j \in E: i \leq j} \rsqrt{w(i)} \left(\vec B(u, i)^* \vec B(u, j)^*\frac{ x_i x_j^*}{\sqrt{\delta(i)} \sqrt{\delta(j)}} + \vec B(u, j)^* \vec B(u, i)\frac{x_j x_i^*}{\sqrt{\delta(j)} \sqrt{\delta(i)}} \right) \rsqrt{w(j)} \\
&& = & \sum_{u \in V}  \frac{1}{d(u)} \sum_{i, j \in E: i \leq j} \rsqrt{w(i)} \left( \frac{x_i^* x_i}{\delta(i)} +  \frac{x_j^* x_j}{\delta(j)}  \right) \rsqrt{w(j)}\\
&&  & - \sum_{u \in V}  \frac{1}{d(u)} \sum_{i, j \in E: i \leq j} \rsqrt{w(i)} \left(\vec B(u, i)^* \vec B(u, j)^*\frac{ x_i x_j^*}{\sqrt{\delta(i)} \sqrt{\delta(j)}} + \vec B(u, j)^* \vec B(u, i)\frac{x_j x_i^*}{\sqrt{\delta(j)} \sqrt{\delta(i)}} \right) \rsqrt{w(j)}  \\ 
&& = & \sum_{u \in V}  \frac{1}{d(u)} \sum_{i, j \in E: i \leq j} \rsqrt{w(i)} \left( \frac{x_i^* x_i}{\delta(i)} +  \frac{x_j^* x_j}{\delta(j)}  - \vec B(u, i)^* \vec B(u, j)^*\frac{ x_i x_j^*}{\sqrt{\delta(i)} \sqrt{\delta(j)}} - \vec B(u, j)^* \vec B(u, i)\frac{x_j x_i^*}{\sqrt{\delta(j)} \sqrt{\delta(i)}} \right)  \rsqrt{w(j)} .
%
\end{align*}

\normalsize

We proceed by analyzing the three possible cases for the summand.

Case 1.a: $u \in H(i) \cap H(j)$ $\Leftrightarrow$ $\vec B(u, i)=1, \vec B(u, j)=1$. We have $\vec B(u, i)^*\vec B(u, j) = \vec B(u, j)^*\vec B(u, i) = 1$.

Case 1.b: $u \in T(i) \cap T(j)$ $\Leftrightarrow$ $\vec B(u, e)=-\ii, \vec B(v, e)=-\ii$. We have $\vec B(u, i)^*\vec B(u, j) = \vec B(u, j)^*\vec B(u, i) = (-\ii)^*(-\ii) = (-\ii)(\ii) = 1$.

In both cases, we have:
\begin{align*}
& \rsqrt{w(i)} \left( \frac{x_i^* x_i}{\delta(i)} +  \frac{x_j^* x_j}{\delta(j)}  - \frac{ x_i x_j^*}{\sqrt{\delta(i)} \sqrt{\delta(j)}} + \frac{x_j x_i^*}{\sqrt{\delta(j)} \sqrt{\delta(i)}} \right)  \rsqrt{w(j)} = \\
& \rsqrt{w(i)} \left(\frac{x_i}{\sqrt{\delta(i)}} - \frac{x_j}{\sqrt{\delta(j)}}\right)^* \left(\frac{x_i}{\sqrt{\delta(i)}} - \frac{x_j}{\sqrt{\delta(j)}}\right) \rsqrt{w(j)}.    
\end{align*}

Letting $x_i = a_i + \ii b_i$ and $x_j = a_j + \ii b_j$, we have:
$$
\rsqrt{w(i)} \left(\left(\frac{a_i}{\sqrt{\delta(i)}} - \frac{a_j}{\sqrt{\delta(j)}} \right)^2 + \left(\frac{b_i}{\sqrt{\delta(i)}} - \frac{b_j}{\sqrt{\delta(j)}}\right)^2\right) \rsqrt{w(j)}.
$$

Case 2.a: $u\in H(i) \cap T(j)$ $\Leftrightarrow$ $\bar B(u, i)=1, \bar B(u, j) = -\ii$. We have $\bar B(u, i)^*  \bar B(u, j) = (1)^*(-\ii) = -\ii$ and $\bar B(u, j)^*  \bar B(u, i) = (-\ii)^*(1) = \ii$.

Thus:

\begin{align*}
& \rsqrt{w(i)} \left( \frac{x_i^* x_i}{\delta(i)} +  \frac{x_j^* x_j}{\delta(j)}  + \ii \frac{ x_i x_j^*}{\sqrt{\delta(i)} \sqrt{\delta(j)}} - \ii \frac{x_j x_i^*}{\sqrt{\delta(j)} \sqrt{\delta(i)}} \right)  \rsqrt{w(j)} \\
\end{align*}
Let $x_i = a_i + \ii b_i$ and $x_j = a_j + \ii b_j$, then we have:
$$
 \rsqrt{w(i)} \left(\left(\frac{a_i}{\sqrt{\delta(i)}} - \frac{b_j}{\sqrt{\delta(j)}} \right)^2 + \left(\frac{a_j}{\sqrt{\delta(j)}} + \frac{b_i}{\sqrt{\delta(i)}}\right)^2\right)\rsqrt{w(j)}.
$$

Case 2.b: $u\in T(i) \cap H(j)$ $\Leftrightarrow$ $\bar B(u, i)=-\ii, \bar B(u, j) = 1$. We have $\bar B(u, i)^*  \bar B(u, j) = (-\ii)^*(1) = \ii$ and $\bar B(u, j)^* \bar B(u, i) = (1)^*(-\ii) = - \ii$.
We have:

\begin{align*}
& \rsqrt{w(i)} \left( \frac{x_i^* x_i}{\delta(i)} +  \frac{x_j^* x_j}{\delta(j)}  - \ii \frac{ x_i x_j^*}{\sqrt{\delta(i)} \sqrt{\delta(j)}} -+ \ii \frac{x_j x_i^*}{\sqrt{\delta(j)} \sqrt{\delta(i)}} \right)  \rsqrt{w(j)} \\
\end{align*}

Let $x_i = a_i + \ii b_i$ and $x_j = a_j + \ii b_j$, then we have:
$$
\rsqrt{w(i)}  \left( \left(\frac{a_i}{\sqrt{\delta(i)}} + \frac{b_j}{\sqrt{\delta(j)}} \right)^2 + \left(\frac{a_j}{\sqrt{\delta(j)}} - \frac{b_i}{\sqrt{\delta(i)}}\right)^2\right) \rsqrt{w(i)} .
$$

The final equation reported in the statement of the theorem is obtained by combining the four cases we just analyzed.
\end{proof}

\Paste{thm:psdL}
\begin{proof}
    Since $\mathbb{\vec L}_N$ is Hermitian, it can be diagonalized as $U \Lambda U^*$ for some $U \in \mathbb{C}^{n \times n}$ and $\Lambda \in \mathbb{R}^{n \times n}$, where $\Lambda$ is diagonal and real.
    We have $x^* \mathbb{\vec L}_N x = x^* U \Lambda U^* x = y^* \Lambda y$ with $y = U^* x$. Since $\Lambda$ is diagonal, we have $y^* \Lambda y = \sum_{u \in V} \lambda_u y_u^2$.
    Thanks to Theorem~\ref{thm:dirichlet}, the quadratic form $x^* \mathbb{\vec L}_N x$ associated with $\mathbb{\vec L}_N$ is a sum of squares and, hence, nonnegative. Combined with $x^* \mathbb{\vec L}_N x = \sum_{u \in L(V)} \lambda_u y_u^2$, we deduce $\lambda_u \geq 0$ for all $u \in L(V)$, where $L(V)$ is the vertex set of DLG($\vec H$).
\end{proof}

\Paste{thm:eigenvalues}
\begin{proof}
$\lambda_{\text{max}}(\mathbb{\vec L}_N) \leq 1$ holds if and only if $\mathbb{\vec L}_N - I \preceq 0$. Since $\mathbb{\vec L}_N = I - \mathbb{\vec Q}_N$ holds by definition, we need to prove $- \mathbb{\vec Q}_N \preceq 0$. This is the case due to Theorem~\ref{thm:psd}.

Similarly, $\lambda_{\text{max}}(\mathbb{\vec Q}_N) \leq 1$ holds if and only if $\mathbb{\vec Q}_N - I \preceq 0$. Since $\mathbb{\vec Q}_N = I - \mathbb{\vec L}_N$ holds by definition, we need to prove $- \mathbb{\vec L}_N \preceq 0$. This is the case due to Theorem~\ref{thm:psdL}.
\end{proof}

\paragraph{\laplaciano{} and The Other Laplacians}

Examining the behavior of the \laplaciano{} through Equation~\ref{eq:laplacian-expanded}, we observe that it differs from other Laplacians designed to handle both directed and undirected edges in graphs, such as the \textit{Magnetic Laplacian}~\citep{lieb1993fluxes} and the \textit{Sign Magnetic Laplacian}~\citep{fiorini2023sigmanet}.
Indeed, the \mbox{\laplaciano{}} exhibits a unique characteristic: both its real and imaginary components can be simultaneously non-zero.
%
This is different from the case of the \textit{Sign Magnetic Laplacian}, which can only have one of the two components different from zero at any given time, and also from the case of the \textit{Magnetic Laplacian}, which coincides with the \textit{Sign Magnetic Laplacian} when $q = \frac{1}{4}$ and the graph has binary weights.
Let us note that the \textit{Magnetic Laplacian} can also have both components different from zero, but such a behavior is influenced by both the edge weight and the value of $q$, and may lead to the sign-pattern inconsistency described in~\cite{fiorini2023sigmanet}, which our proposed \mbox{\laplaciano{}} does not suffer from.
%

\section{Complexity of \rete{}}\label{appx:complexity}

The detailed calculations for the (inference) complexity of \rete{} are as follows.
%
\begin{enumerate}
    \item The \laplaciano{} $\mathbb{\vec L}_N$ is constructed in time $O(m^2 n)$, where the factor $n$ is due to the need for computing the product between two columns of $\vec B$ (i.e., two rows of $B^*$) to calculate each entry of $\mathbb{\vec L}_N$. After $\mathbb{\vec L_N}$ has been computed, the convolution matrix $\hat Y \in \mathbb{C}^{m \times m}$ is constructed in time $O(m^2)$. Note that such a construction is carried out entirely in pre-processing and is not required at inference time.
    \item Constructing the feature matrix $X = \vec{B}^* X'$ requires $O(m n c_0)$ elementary operations.
    \item Each of the $\ell$ convolutional layers of \rete{} requires $O(m^2 c + m c^2 + m c) = O(m^2 c + m c^2)$ elementary operations across 3 steps. Let $X^{l-1}$ be the input matrix to layer $l = 1, \dots, \ell$. The operations that are carried out are the following ones.
    \begin{enumerate}
        \item $\mathbb{\vec L_N}$ is multiplied by the hyperedge-feature matrix $X^{l-1} \in \mathbb{C}^{m \times c}$, obtaining $P^{l_1} \in \mathbb{C}^{m \times c}$ in time $O(m^2 c)$ (we assume, for simplicity, that matrix multiplications takes cubic time);
        \item The matrices $P^{l_0} = I X^{l-1} = X^{l-1}$ and $P^{l1}$ are multiplied by the weight matrices $\Theta_0, \Theta_1 \in \mathbb{R}^{c \times c}$ (respectively), obtaining the intermediate matrices $P^{l_{01}}, P^{l_{11}} \in \mathbb{C}^{n \times c}$ in time $O(mc^2)$ .
        \item The matrices $P^{l_{01}}$ and $P^{l_{11}}$ are additioned in time $O(m c)$ to obtain $P^{l_2}$.
        \item The activation function $\phi$ is applied component wise to $P^{l_2}$ in time $O(m c)$, resulting in the output matrix $X^{l} \in \mathbb{C}^{m \times c}$ of the $l$-th convolutional layer.
    \end{enumerate}
    %
    \item The unwind operator transforms $X^{\ell}$ (the output of the last convolutional layer $\ell$) into the matrix $U^{0} \in \mathbb{R}^{n \times 2 c}$ in linear time $O(m c)$.
    \item Call $U^{s-1}$ the input matrix to each linear layer of index $s = 1, \dots, S$. The application of the $s$-th linear layer to $U^{s-1} \in \mathbb{C}^{m \times c'}$ requires multiplying $U^{s-1}$ by a weight matrix $M_s \in \mathbb{C}^{c' \times c'}$
    (where $c'$ is the number of channels from which and into which the feature vector of each node is projected). This is done in time $O(m c'^2)$. 
    \item In the last linear layer of index $S$, the input matrix $U^{S-1} \in \mathbb{R}^{m \times c'}$ is projected into the output matrix $O \in \mathbb{R}^{m \times d}$ in time $O(n c' d)$.
    \item The application of the Softmax activation function takes linear time $O(m d)$.
\end{enumerate}

We deduce an overall complexity of
$
O(m n c_0) 
+ O(\ell (m^2 c + m c^2) + mc  + (S-1)(m c'^2) + m c' d + md)$.
Assuming $O(c) = O(c') = O(d) = \bar c$, 
such a complexity coincides with
$O(\ell (m^2 \bar c) + (\ell + S) (m \bar c^2))$.

\section{Further Details on the Datasets}\label{appx:dataset}
Details on the datasets composition are reported in Tables~\ref{tab:uspto},~\ref{tab:ord},~\ref{tab:sn2}. 
Most of the elements of {\tt Dataset-1} belong to the first two classes, which concern the addition of functional groups to a chemical compound: alkyl and aryl groups for \textit{Class 1} and acyl groups for \textit{Class 2}, comprising more than 17K species. Less populated classes involve specific chemical transformations, such as \textit{Class 3} (C--C bond formation) which contains less than 1000 elements. 

{\tt Dataset-2} presents solely three classes; nevertheless, we have been able to extract sub-categories from two of them. Those are \textit{Class 1} (C--C bond formation) and \textit{Class 2} (N-arylation processes) and contain two and three sub-classes, respectively. The most populated class is Imidazole-aryl coupling, comprising around 1500 elements belonging to the class of palladium-catalyzed imidazole C-H arylation. The chemical diversity in this class is ensured by the use of 8 aryl bromides and 8 imidazole compunds. Furthermore, in terms of reaction conditions, the collection presents 24 different monophosphine ligands.         

Unlike the previous ones, {\tt Dataset-3} has been assembled starting from competitive processes; therefore it contains almost the same amount of elements ($\sim$ 300) for the two classes: Bimolecular nucleophilic substitution (S$\mathrm{_N}$2) and eliminations (E2).
The reactants--which are in common between S$\mathrm{_N}$2 and E2---are substituted alkane compounds and nucleophile agents. The substituents span a range of electron donating and electron withdrawing effect strengths, including methyl, cyano, amine, and nitro functional groups. The nucleophiles have been chosen either between halide or hydrogen anions, while the molecular skeleton is ethane.

\begin{table}[htb!]
 \centering
\caption{Distribution of the reactions in the  Dataset-1. }
\resizebox{.7\columnwidth}{!}{%
\footnotesize
\begin{tabular}{lll}    
     Rxn class & Rxn name & Num rxns \\
    \hline
        1 &  Heteroatom alkylation and arylation & 15151\\
2 & Acylation and related process & 11896\\
3 &  C-C bond formation & 909\\
4& Heterocycle formation & 4614\\
5& Protections & 1834\\
6& Deprotections & 5662\\
7& Reductions & 672\\
8& Oxidations & 811\\
9& Functional group interconversion & 8237\\
10 & Functional group addition & 230\\
\hline
    \end{tabular}
    \label{tab:uspto}
    }
\end{table}

 \begin{table}[htb!]
\centering
\caption{Distribution of the reactions in the  Dataset-2. }
\resizebox{.7\columnwidth}{!}{%
\begin{tabular}{lll}    
     
     Rxn class & Rxn name & Num rxns \\
    \hline
   
       1  & C-C bond formation & 1921 \\
          & - Reizman Suzuki Cross-Coupling & 385\\
          &  - Imidazole-aryl coupling & 1536\\
        2 & Heteroatom (N) arylation: & 657 \\
        & - Amine  + Aryl bromide   & 278 \\
        & - Amine + Aryl chloride  & 299 \\
        & - Amine + Aryl iodide  & 80 \\
         3 & Amide bond formation & 960  \\
\hline       
    \end{tabular}
    \label{tab:ord}
    }
\end{table}

 \begin{table}[htb!]
\centering
\caption{Distribution of reactions in the  Dataset-3. }
\resizebox{.8\columnwidth}{!}{%
\begin{tabular}{lll}    
    Rxn class & Rxn name & Num rxns \\
    \hline
       1  & Bimolecular nucleophilic substitution (S$\mathrm{_N}$2) & 301 \\
        2 & Bimolecular elimination (E2) & 348\\
        \hline
    \end{tabular}
    \label{tab:sn2}
  }  
\end{table}

\section{Further Details on the Experiments}\label{appx:experiment}

\paragraph{Hardware.} The experiments were conducted on 2 different machines: 
\begin{enumerate}
    \item An Intel(R) Xeon(R) Gold 6326 CPU @ 2.90GHz with 380 GB RAM, equipped with an NVIDIA Ampere A100 40GB.
    \item A 12th Gen Intel(R) Core(TM) i9-12900KF CPU @ 3.20GHz CPU with 64 GB RAM, equipped with an NVIDIA RTX 4090 GPU.
\end{enumerate}

\paragraph{Model Settings.} We trained every learning model considered in this paper for up to 1000 epochs. We adopted a learning rate of $5 \cdot 10^{-3}$ and employed the optimization algorithm Adam with weight decays
equal to $5 \cdot 10^{-4}$ (in order to avoid overfitting). We set the number of linear layers to 2, i.e. $\ell = 2$, for all the models. 

We adopted a hyperparameter optimization procedure to identify the best set of parameters for each model. In particular, the hyperparameter values are:
\begin{itemize}
    \item For AllDeepSets and ED-HNN, the number of basic block is chosen in $\{1, 2, 4, 8\}$, the number of MLPs per block in $\{1, 2\}$, the dimension of the hidden MLP (i.e., the number of filters) in $\{64, 128, 256, 512\}$, and the classifier hidden dimension in $\{ 64, 128, 256\}$.
    \item For AllSetTransformer the number of basic block is chosen in $\{2, 4, 8\}$, the number of MLPs per block in $\{1, 2\}$, the dimension of the hidden MLP in $\{64, 128, 256, 512\}$, the classifier hidden dimension in $\{ 64, 128, 256\}$, and the number of heads in $\{ 1, 4, 8\}$.
    \item For UniGCNII, HGNN, HNHN, HCHA/HGNN$^+$, LEGCN, and HCHA with the attention mechanism, the number of basic blocks is chosen in $\{2, 4, 8\}$ and the hidden dimension of the MLP layer in $\{64, 128, 256, 512\}$.
    \item For HyperGCN, the number of basic blocks is chosen in $\{2, 4, 8\}$.
    \item For HyperND, the classifier hidden dimension is chosen in $\{ 64, 128, 256\}$.
    \item For PhenomNN, the number of basic blocks is chosen in $\{2, 4, 8\}$. We select four different settings: 
    \begin{enumerate}
        \item $\lambda_0 = 0.1$, $\lambda_1=0.1$ and prop step$=8$,
        \item $\lambda_0 = 0$, $\lambda_1=50$ and prop step$=16$,
        \item $\lambda_0 = 1$, $\lambda_1=1$ and prop step$=16$, 
        \item  $\lambda_0 = 0$, $\lambda_1=20$ and prop step$=16$.
    \end{enumerate}
    \item For DHM, the number of basic blocks is chosen in $\{1, 2, 3, 4\}$ and the classifier hidden dimension is chosen in $\{ 64, 128, 256, 512\}$.
    \item For \rete{},
    the number of convolutional layers is chosen in $\{1, 2, 3\}$, the number of filters in $\{64, 128, 256, 512\}$, and the classifier hidden dimension in $\{ 64, 128, 256\}$. We tested \rete{} both with the input feature matrix $X \in \mathbb{C}^{n \times c}$ where $\real(X) = \imaginary(X) \neq 0$ and with $\imaginary(X) = 0$.
\end{itemize}

\paragraph{How to Transfer The Features.}\label{appx:features}


%
%

As mention in Section~\ref{sec:rete}, a key aspect of our approach  involves transferring features from the nodes of the hypergraph to their corresponding hyperedges, i.e., the nodes of the directed line graph. To clarify this mechanism, we provide a simple example.
Consider a directed hypergraph $\vec H = (V, E)$, where the vertex set is $V = \{u, v, c\}$ and the hyperedge set consists of $E = \{e_1\}$. 
In $\vec H$, we have $H(e_1) = \{u, v\}$ and $T(e_1) = \{c\}$.
Each vertex is assigned a feature vector $x'_u, x'_v, x'_c = 1$ and the hyperedge has a unit weight, i.e. $w_{e_{1}} = 1$. 
Recalling that  $X = B^{*}X'$, the feature vector $x_1$ of the hyperedge $e_1$ is then calculated as:
$$
x_1 = \vec B_{1u}^* \cdot x_u + \vec B_{1v}^* \cdot x_v  + \vec B_{1c}^* \cdot x_c  = 2 + i.
$$
In the case where $\vec{H} = H$, i.e., when the hypergraph is undirected, we have $\vec{B}^* = B^\top$.
The feature vector $x_1$ of the hyperedge $e_1$ is then calculated as:
$$
x_1 = B_{1u} \cdot x_u + B_{1v} \cdot x_v  + B_{1c} \cdot x_c = 3.
$$
As illustrated by this example , in the specific case of a directed line graph, the feature vector can feature both real and imaginary components, depending on the topology of the hypergraph encoded by $\vec{B}$.

\section{Confusion Matrix}\label{appx:confusionmatrix}

\begin{figure}[htb!]
    \centering
    \includegraphics[width=0.9\linewidth]{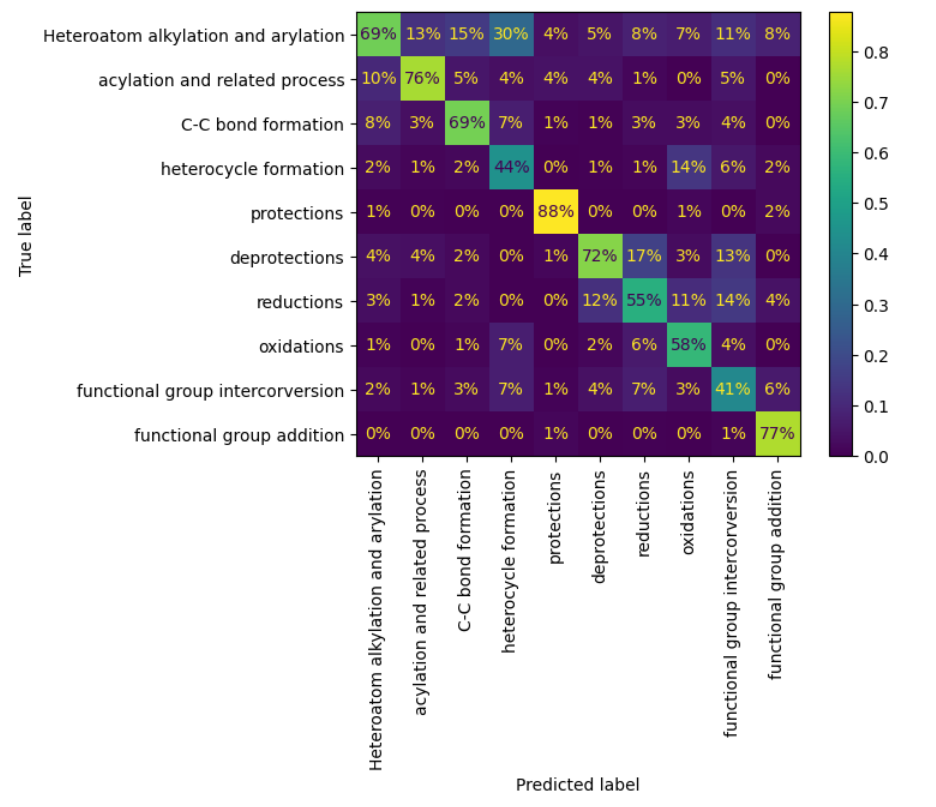}
    \caption{{\tt Dataset-1} confusion matrix.}
    \label{fig:dataset-1}
\end{figure}

\begin{figure}[htb!]
    \centering
    \includegraphics[width=0.7\linewidth]{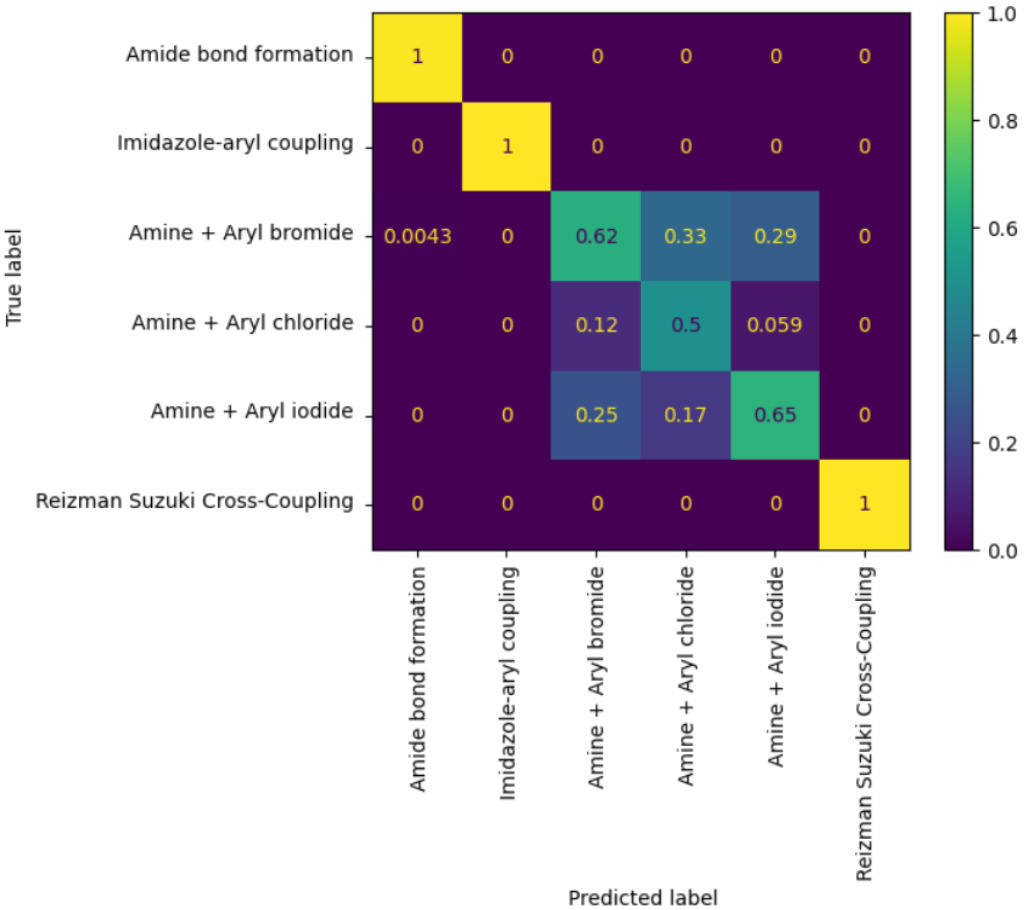}
    \caption{{\tt Dataset-2} confusion matrix.}
    \label{fig:dataset-2}
\end{figure}

We report the confusion matrices of {\tt Dataset-1} in Figure~\ref{fig:dataset-1} and {\tt Dataset-2} in Figure~\ref{fig:dataset-2}. We can extract some insights from these two matrices, in particular:
\begin{itemize}
    \item {\tt Dataset-1.} \rete{} achieves a maximum performance of 88\% in classifying the \textit{Class 5} (Protections). However, its performance drops for \textit{Class 4} and \textit{Class 9} (Heterocycle formations and Functional group interconversions), where it correctly predicts only 44\% and 41\%, respectively.
    \item {\tt Dataset-2.} \rete{} accurately classifies the sub-classes relative to the C--C bond formations (Reizman Suzuki Cross-Coupling and Imidazole-aryl coupling), as well as the Amide bond formations.   
    On the other hand, the remaining three N-arylation sub-classes  are poorly discriminated.
    This behavior can likely be attributed to the fact that the former are derived from different collections of Pd-catalyzed cross-coupling reactions, 
    each exhibiting distinct features in terms of participant molecules (e.g. imidazole compounds). In contrast, 
     all of the elements in the N-arylation classes share the same reaction mechanism (Buchwald-Hartwig amination); this poses a greater challenge, which results in decreased accuracy when predicting the correct class.
\end{itemize}

\section{From a Directed Hypergraph to the Directed Line Graph Laplacian}\label{appx:complex_laplacian}

\begin{figure}[h]
    \centering
    \includegraphics[width=0.5\linewidth]{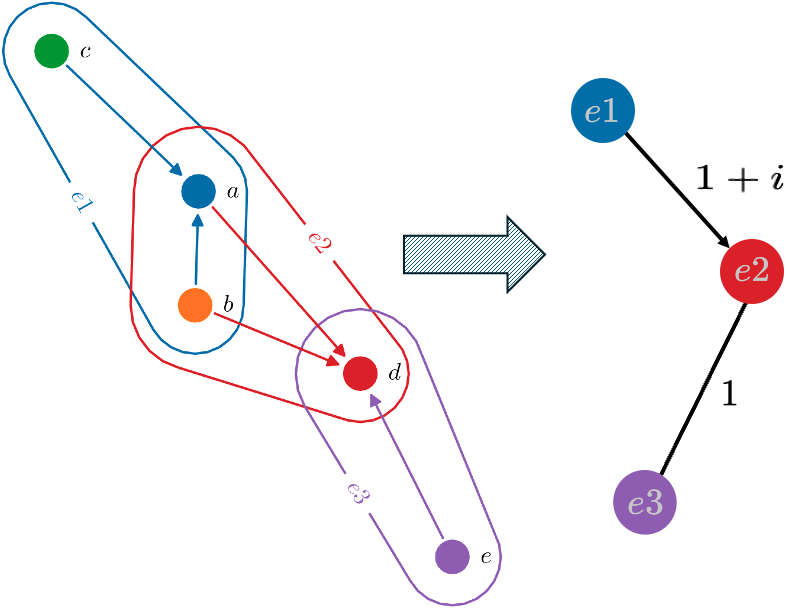}
    \caption{An example illustrating the transformation of a hypergraph (left) into its corresponding directed line graph (right).}
    \label{fig:example_DLG}
\end{figure}

    





To illustrate the construction of the directed line graph and the associated \laplaciano{}, consider a directed hypergraph $\vec H = (V, E)$ where the vertex set is $V = \{a, b, c, d, e\}$ and the hyperedge set is $E = \{e_1, e_2, e_3\}$. The incidence relationships are defined as follows: 
\begin{itemize}
    \item $H(e_1) = \{ b, c\}$, $T(e_1) = \{a\}$,
    \item  $H(e_2) = \{a, b\}$, $T(e_2) = \{d\}$, 
    \item  $H(e_3) = \{e\}$, $T(e_3) = \{d\}$. 
\end{itemize}

Each hyperedge is assigned a unit weight (i.e., $W = I$). The cardinalities (densities) of the hyperedges are $\delta_{e_1} = 3$, $\delta_{e_2} = 2$, and $\delta_{e_3} = 2$.

We construct $DLG(\vec H)$ using the following matrices: the incidence matrix $\vec{B}$, its conjugate transpose $\vec{B}^*$, the vertex degree matrix $D_v$, and the hyperedge degree matrix $D_e$. The incidence matrix $\vec{B}$ and its conjugate transpose are:

\[
\vec B = \begin{bmatrix}
    -i  & 1 & 0 \\
    1   & 1 & 0 \\
    1   & 0 & 0 \\
    0   & -i & -i \\
    0   & 0  & 1
\end{bmatrix}
\quad
\vec B^* = \begin{bmatrix}
    i  & 1 & 1 & 0 & 0 \\
    1  & 1 & 0 & i & 0 \\
    0  & 0 & 0 & i & 1
\end{bmatrix}.
\]

The vertex degree matrix $D_v$ and the hyperedge degree matrix $D_e$ are given by:

\[
D_v = \begin{bmatrix}
    2 & 0 & 0 & 0 & 0 \\
    0 & 2 & 0 & 0 & 0 \\
    0 & 0 & 1 & 0 & 0 \\
    0 & 0 & 0 & 2 & 0 \\
    0 & 0 & 0 & 0 & 1 \\
\end{bmatrix}
\quad
D_e = \begin{bmatrix}
    3 & 0 & 0 \\
    0 & 3 & 0 \\
    0 & 0 & 2
\end{bmatrix}.
\]

Using these matrices, the adjacency matrix $A$ of the directed line graph $DLG(\vec{H})$ is:

\begin{equation}\label{eq:example_adj}
A = \vec{B}^* \vec{B} - D_e = \begin{bmatrix}
    0 & 1+i  & 0 \\
    1-i & 0  & 1 \\
    0 & 1 & 0 
\end{bmatrix}.    
\end{equation}

By Definition \ref{def:DLG}, the directed line graph $DLG(\vec H)$ has three vertices, corresponding to the hyperedges $e_1$, $e_2$, and $e_3$ of the original hypergraph $\vec H$. An edge exists between two vertices in $DLG(\vec H)$ if and only if their corresponding hyperedges in $\vec H$ are incident. 
In the specific example (illustrated in Figure \ref{fig:example_DLG}), DLG$(\vec H)$ contains two edges, whose direction and weight are determined by the adjacency matrix $A$ (in equation \ref{eq:example_adj}. 
Without loss of generality, we consider the upper triangular part of $A$ to assign weights to the edges and define the directions: In the example considered, one edge will be directed and have a weight equal to $1+i$ (i.e. $e_1 \overset{1+i}{\rightarrow} e_2$), the other edge will be undirected and have a weight equal to 1 ($e_2 \;\overset{1}{ \rule[0.5ex]{0.3cm}{0.5pt} }\; e_3$).

Using the equation \ref{eq:myproposal2}. we can calculate the proposed \laplaciano{} $\mathbb{\vec L}_N$ as follows:

\[
\mathbb{\vec L}_N = I - \mathbb{\vec{Q}}_{N} :=  \rsqrt{\vec{D}_e} \vec{B}^* \vec{D}_v^{-1} \vec{B} \rsqrt{\vec{D}_e} = 
\begin{bmatrix}
0.333 & -0.167 -167 i & 0 \\
-0.167+ 0.167 i & 0.5 & -0.204 \\
0 & -0.204 & 0.25    
\end{bmatrix}.
\]

By inspecting $\mathbb{\vec{L}_N}$, one can observe that it encodes the elements of the hypergraph $\vec H$ in the following way:
\begin{enumerate}
    \item The real components of off-diagonal entries in $\mathbb{\vec{L}_N}$ encode the fact that, in the underlying hypergraph $\vec H$, the vertex belongs to the head set or tail set simultaneously in two different hyperedges. For example, $\mathbb{\vec{L}_N}(2,3) = -0.204$ indicates that $H(e_2) \cap H(e_3) \neq \emptyset$ or $T(e_2) \cap T(e_3) \neq \emptyset$. In this specific case, $T(e_2) \cap T(e_3) = \{d\}$. Similarly, $\real(\mathbb{\vec{L}_N}(1,2)) = -0.167$ arises from the fact that $e_1$ and $e_2$ share the vertex $b$ in their head sets.

    \item The imaginary component captures the hyperedge directionality based on the underlying hypergraph $\vec{H}$, where a node belongs to the head set of one hyperedge and the tail set of another. For example, $\imaginary(\mathbb{\vec{L}}_N(1,2)) = -\imaginary(\mathbb{\vec{L}}_N(2,1)) = -0.167$, indicating that $a \in T(e_1) \cap H(e_2)$. 
    \item The absence of any relationships between hyperedges $e1$ and $e3$ is encoded by $0$ in $DGL(\vec H)$. 
    Specifically, $\mathbb{\vec L}_N(1,3) = \mathbb{\vec L}_N(3,1) = 0$.
    \item The \textit{self-loop information} (a measure of how strongly the feature of a vertex depends on its current value within the convolution operator) is encoded by the diagonal of $\mathbb{\vec L}_N$.
\end{enumerate}

\end{document}